%% file: egpaper_for_review.tex
\documentclass[10pt,twocolumn,letterpaper]{article}

\usepackage{iccv}
\usepackage{times}
\usepackage{epsfig}
\usepackage{graphicx}
\usepackage{amsmath}
\usepackage{amssymb}
\usepackage{bm}
\usepackage{multirow}
\usepackage{booktabs}
\usepackage{pifont}

\usepackage{algorithm}
\usepackage{algorithmic}
\usepackage{caption}
\usepackage{subcaption}

\def\mbfx{\mathbf{x}}
\def\mdata{\mathrm{data}}
\def\mmean{\mathbb{E}}

\def\CELEHQRES{7.75}
\def\CHURCHRES{8.97}
\def\IMGNETACC{72.5}
\def\IMGNETIS{189.5}
\def\IMGNETFID{6.77}

\def\IMGNETWAACC{78.9}
\def\IMGNETWAIS{231.3}
\def\IMGNETWAFID{6.05}

\def\OURNAME{EGC}
\def\UCNAME{Unsupervised EGC model}

\usepackage[pagebackref=true,breaklinks=true,letterpaper=true,colorlinks,bookmarks=false]{hyperref}

\iccvfinalcopy 

\def\iccvPaperID{****} 

\ificcvfinal\fi

\begin{document}

\title{EGC: Image Generation and Classification via \\a Diffusion Energy-Based Model }

\author{Qiushan Guo\textsuperscript{1}, Chuofan Ma\textsuperscript{1}, Yi Jiang\textsuperscript{2}, Zehuan Yuan\textsuperscript{2}, Yizhou Yu\textsuperscript{1}, Ping Luo\textsuperscript{1}\\
\textsuperscript{1}The University of Hong Kong
\textsuperscript{2}ByteDance Inc. \\
}

\twocolumn[{%
\renewcommand\twocolumn[1][]{#1}%
\maketitle
\begin{center}
  \centering
  \captionsetup{type=figure}
  \includegraphics[width=\linewidth]{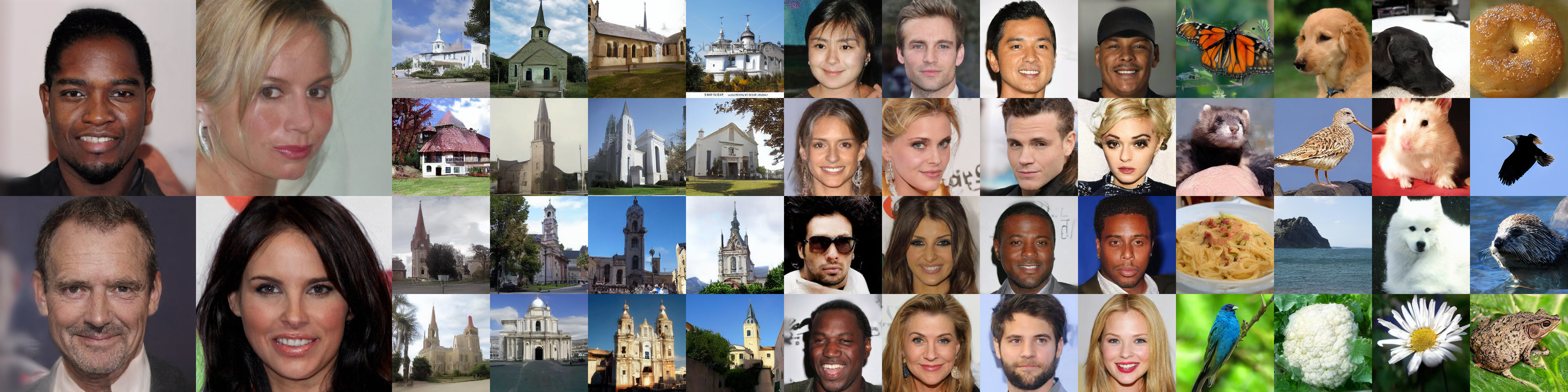}
  \captionof{figure}{
    Generated samples from our \OURNAME\ models on CelebA-HQ 1024$\times$1024, LSUN Church 256$\times$256, CelebA-HQ 256$\times$256 and ImageNet 256$\times$256 datasets, shown from left to right.
  }
\end{center}\label{fig:CIFAR10}
}]

\ificcvfinal\thispagestyle{empty}\fi

\begin{abstract}
\vspace{-2pt}
Learning image classification and image generation using the same set of network parameters is a challenging problem. 
Recent advanced approaches perform well in one task often exhibit poor performance in the other. 
This work introduces an energy-based classifier and generator, namely \OURNAME , which can achieve superior performance in both tasks using a single neural network.
Unlike a conventional classifier that outputs a label given an image (\ie, a conditional distribution $p(y|\mathbf{x})$), the forward pass in \OURNAME\ is a classifier that outputs a joint distribution $p(\mathbf{x},y)$, enabling an image generator in its backward pass by marginalizing out the label $y$. 
This is done by estimating the classification probability given a noisy image from the diffusion process in the forward pass, while denoising it using the score function estimated in the backward pass.
\OURNAME\  achieves competitive generation results compared with state-of-the-art approaches on ImageNet-1k, CelebA-HQ and LSUN Church, while achieving superior classification accuracy and robustness against adversarial attacks on CIFAR-10. 
This work represents the first successful attempt to simultaneously excel in both tasks using a single set of network parameters.
We believe that \OURNAME\ bridges the gap between discriminative and generative learning. Code will be released at \href{https://github.com/GuoQiushan/EGC}{https://github.com/GuoQiushan/EGC}.
\end{abstract}


\begin{figure}
\vspace{20pt}
\begin{center}
    \includegraphics[trim=0.6cm 0.3cm 0.1cm -0.1cm,width=\columnwidth]{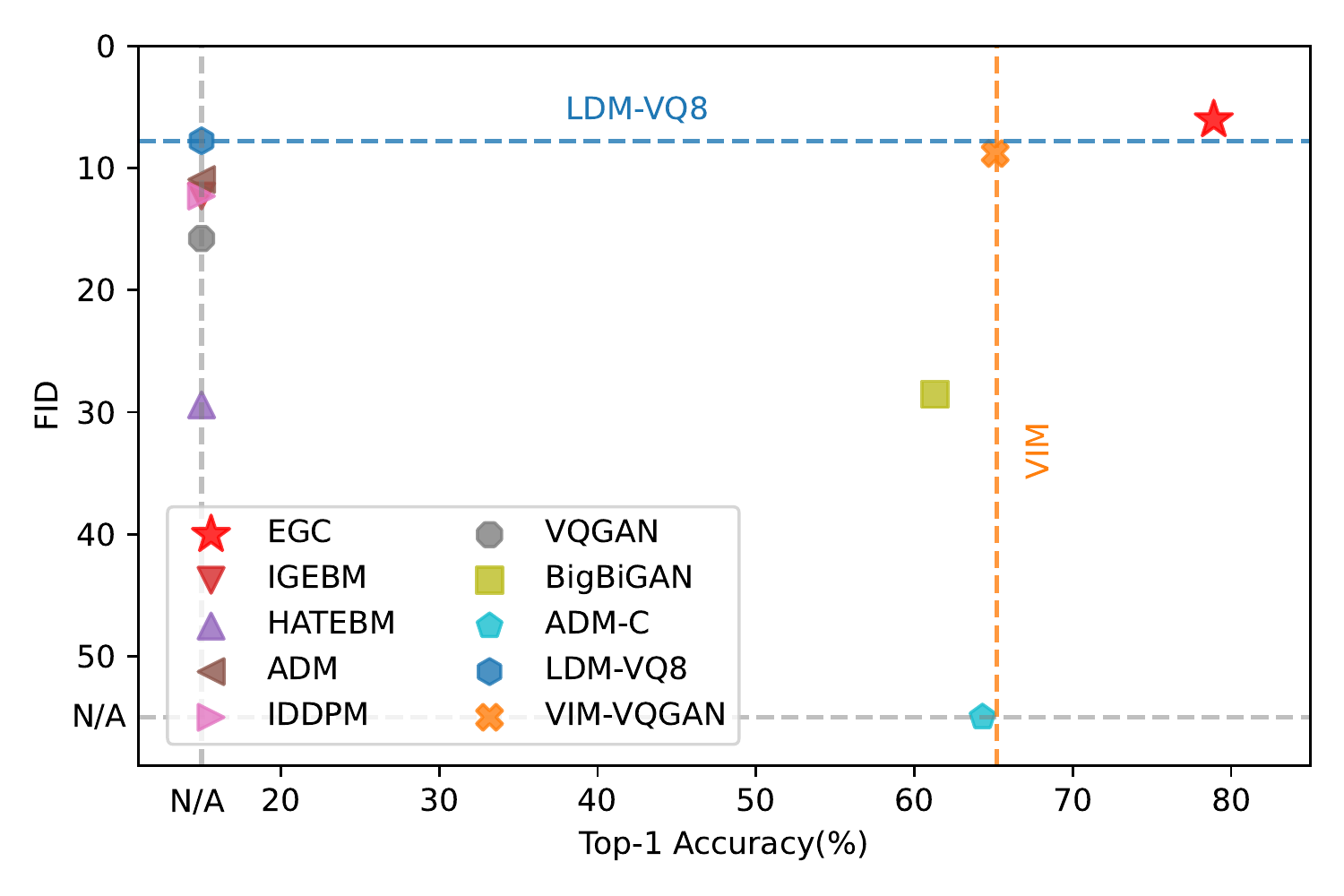}
\end{center}
\vspace{-8pt}
\caption{FID and classification accuracy on ImageNet 256$\times$256 dataset. The scatters plots on the vertical N/A line represent the image generation models, which are not available for classification. The scatter plot on the horizontal N/A line represents the classification model, which is not available for image generation. Remarkably, \OURNAME\  achieves superior performance in both tasks with a single neural network, demonstrating its effectiveness in bridging the gap between discriminative and generative learning.}
\label{Fig:Scatter}
\end{figure}

\input{intro.tex}

\input{relate.tex}
\input{method.tex}

\input{exp.tex}

\input{appendix.tex}

{\small
\bibliographystyle{ieee_fullname}
\bibliography{egbib}
}

\end{document}

%% file: intro.tex
\section{Introduction}

Image classification and generation are two fundamental tasks in computer vision that have seen significant advancements with the development of deep learning models. 
However, many state-of-the-art approaches that perform well in one task often exhibit poor performance in the other or are not suitable for the other task.
Both tasks can be formulated from a probabilistic perspective, where image classification task is interpreted as a conditional probability distribution $p(y|\mbfx)$, and image generation task is the transformation of a known and easy-to-sample probability distribution $p(\mathbf{z})$ to a target distribution $p(\mathbf{x})$.

As an appealing class of probabilistic models, 
Energy-Based Models (EBM) \cite{du2019implicit,finn2016connection,gao2018learning,gao2020flow,gao2020learning,goyal2017variational,grathwohl2019your,kim2016deep,kumar2019maximum,neal2011mcmc,nijkamp2019learning,wang2023energy,xie2016theory,yin2022learning,zhao2016energy} can explicitly model complex probability distribution and be trained in an unsupervised manner. Furthermore, standard image classification models and some image generation models can be reinterpreted as an energy-based model \cite{gao2020learning,grathwohl2019your, jin2017introspective,lazarow2017introspective,lee2018wasserstein,liu2020energy}. From the perspective of EBM, a standard image classification model can be repurposed as an image generation model by leveraging the gradient of its inputs to guide the generation of new images.
Despite the desirable properties, EBMs face challenges in training due to the intractability of computing the exact likelihood and synthesizing exact samples from these models.
Arbitrary energy models often exhibit sharp changes in gradients, leading to unstable sampling with Langevin dynamics. 
To ameliorate this issue, spectral normalization \cite{miyato2018spectral} is typically adopted for constraining the Lipschitz constant of the energy model \cite{du2019implicit,gao2020learning,grathwohl2019your}.
Even with this regularization technique,
the samples generated by the energy-based model are still not competitive enough because the probability distribution of real data is usually sharp in the high-dimensional space, providing inaccurate guidance for image sampling in the low data density regions.

Diffusion models \cite{ho2020denoising,rombach2022high,song2020denoising,song2019generative,song2020improved} have demonstrated competitive and even superior image generation performance compared to GAN \cite{goodfellow2020generative} models. In diffusion models, images are perturbed with Gaussian noise through a diffusion process for training, and the reverse process is learned to transform the Gaussian distribution back to the data distribution. As pointed out in \cite{song2019generative}, perturbing data points with noise populates low data density regions to improve the accuracy of estimated scores, resulting in stable training and image sampling.

Motivated by the flexibility of EBM and the stability of diffusion model, we propose a novel energy-based classifier and generator, namely \OURNAME , which achieves superior performance in both image classification and generation tasks using a single neural network. \OURNAME\  is a classifier in the forward pass and an image generator in the backward pass.
Unlike a conventional classifier that predicts the condition distribution $p(y|\mbfx)$ of the label given an image, the forward pass in \OURNAME\  models the joint distribution $p(\mbfx, y)$ of the noisy image and label, given the source image. 
By marginalizing out the label $y$, the gradient of log-probability of the noisy image (\ie, unconditional score) is used to restore image from noise. The classification probability $p(y|\mbfx)$ provides classifier guidance together with unconditional score within one step backward pass.

We demonstrate the efficacy of \OURNAME\  model on ImageNet, CIFAR-10, CIFAR-100, CelebA-HQ and LSUN datasets. The generated samples are of high fidelity and comparable to GAN-based methods, as shown in Fig.~1. Additionally, our model shows superior classification accuracy and robustness against adversarial attacks.
On CIFAR-10, \OURNAME\ surpasses existing methods of learning explicit EBMs with an FID of 3.30 and an inception score of 9.43 while achieving a remarkable classification accuracy of 95.9\%.
This result even exceeds the classification performance of the discriminative model Wide ResNet-28-12, which shares a comparable architecture and number of parameters with our model.
On ImageNet-1k,\ \OURNAME\ achieves an FID of\ \IMGNETWAFID\ and an accuracy of\ \IMGNETWAACC\%, as illustrated in Fig.~\ref{Fig:Scatter}. 
We also demonstrate that naively optimizing the gradients of explicit energy functions as the score functions outperforms optimizing the probability density function via Langevin sampling. Besides, \OURNAME\  model does not require constraining the Lipschitz constant as in the previous methods \cite{du2019implicit,gao2020learning,grathwohl2019your} by removing the normalization layers and inserting spectral normalization layer. 
More interestingly, we demonstrate that the neural network effectively models the target data distribution even though
we adopt optimization of the Fisher divergence instead of the probability $p_\theta(\mbfx_t)$. 

Our contributions are listed as follows: 

(1) We propose a novel energy-based model, \OURNAME , bridging the gap between discriminative and generative learning. In \OURNAME , the forward pass is a classification model that predicts the joint distribution $p(\mbfx, y)$ and the backward pass is a generation model that denoises data using the score function and conditional guidance.
(2) Our \OURNAME\  model achieves competitive generation results to state-of-the-art approaches, while obtaining superior classification results using a single neural network. \OURNAME\  surpasses existing methods of explicit EBMs by a significant margin.
(3) We demonstrate that \OURNAME\  model can be applied in inpainting, semantic interpolation, high-resolution image generation ($\sim$ 1024$^2$) and robustness improvement.

%% file: relate.tex
\begin{figure*}
\begin{center}
    \includegraphics[width=2\columnwidth]{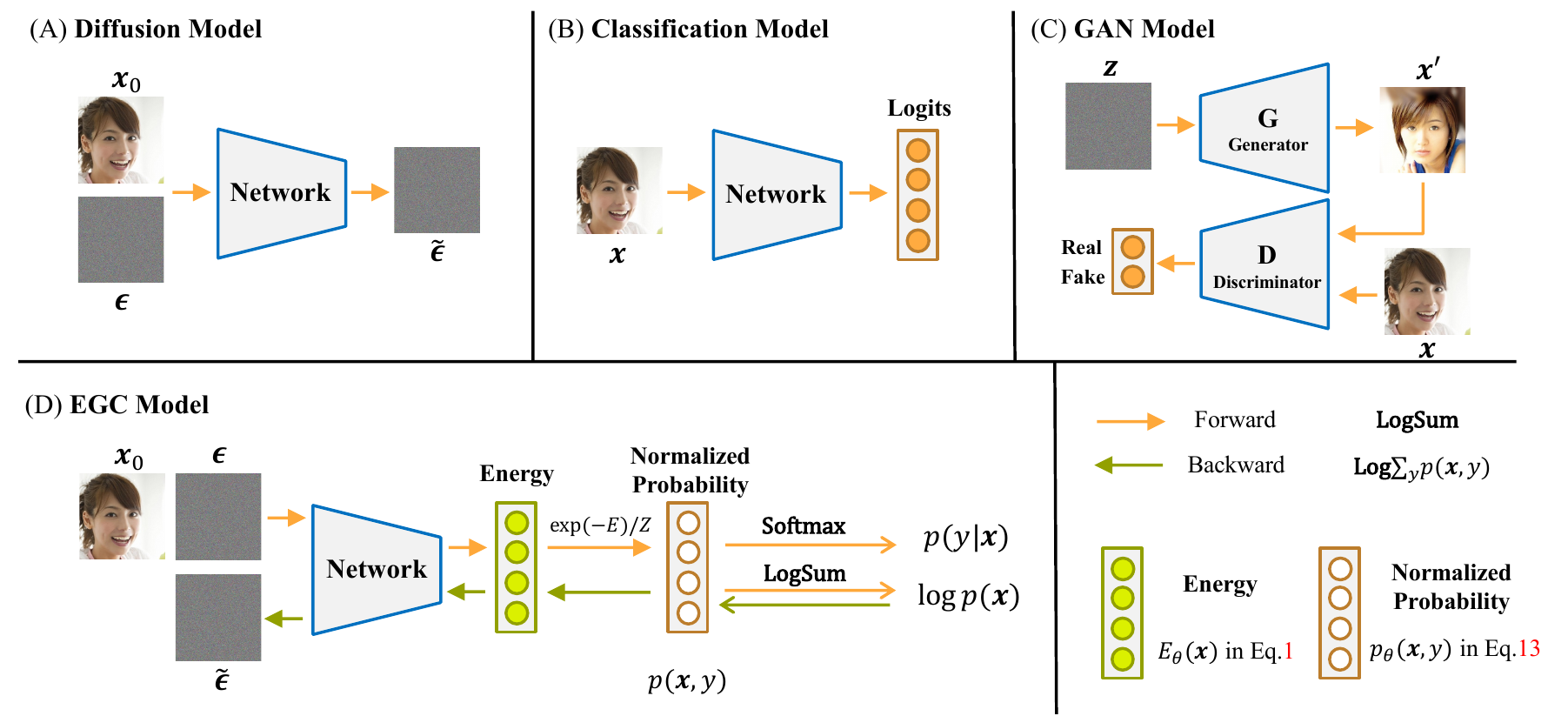}
\end{center}
\vspace{-15pt}
\caption{(A) Diffusion Model estimates the score (noise) from the noisy scaled image. (B) Standard Classification Model outputs the logits for minimizing the cross-entropy loss. (C) GAN Model is composed of a generator model that synthesizes new samples and a discriminator that classifies samples as either real or fake. (D) \OURNAME\  Model estimates the joint distribution $p(\mathbf{x}, y)$ for classification via the forward propagation of a neural network and leverages the score estimated from the backward propagation to generate samples from Gaussian noise. $Z$ represents the normalizing constant, which is only relevant to the model parameters.}
\label{Fig:EBDC}
\vspace{-10pt}
\end{figure*}

\section{Related Works}
\textbf{Energy-based Models.} Unlike most other probabilistic models, Energy-Based Models do not place a restriction on the tractability of the normalizing constant, which confers upon them greater flexibility in modeling complex probability distributions. However, the intractability of the normalization constant renders their training challenging. BiDVL \cite{kan2022bi} proposes a bi-level optimization framework to facilitate learning of energy-based latent variable models. Yin \etal \cite{yin2022learning} explore adversarial training for learning EBMs. CLEL \cite{leeguiding} improves the training of EBMs using contrastive representation learning, which guides EBMs to better understand the data structure for faster and more memory-efficient training. EGSDE \cite{zhao2022egsde} proposes energy-guided stochastic differential equations to guide the inference process of a pretrained SDE for realistic and faithful unpaired image-to-image translation. In contrast to the above methods, we reinterpret the standard classification network as an EBM to approximate the joint distribution of the noisy samples in diffusion process. Our method enables the forward pass to work as a classification model and the backward pass to provide both unconditional and conditional scores within one step. 

\textbf{Denoising Diffusion Models (DDMs)}, originating from \cite{sohl2015deep}, are to learn from noisy data and generate samples by reversing the diffusion process.
Score based generative model \cite{song2019generative} is introduced to train denoising models with multiple noise levels and draw samples via Langevin dynamics during inference. 
Several works have improved the design and demonstrated the capability of synthesizing high-quality images \cite{ho2020denoising, dhariwal2021diffusion, classifier-free}. 
DDPM\cite{ho2020denoising} generates high-quality images synthesis results using diffusion probabilistic models. 
Guided-Diffusion\cite{dhariwal2021diffusion} firstly achieved better performance than GAN by utilizing the classifier guidance. 
Latent Diffusion\cite{rombach2022high} further proposes latent diffusion models which operate on a compressed latent space of reduced dimensionality to save computational cost. DDMs circumvent the issue of intractable normalizing constants in EBMs by modeling the score function instead of the density function. Different from the score-based diffusion models, our \OURNAME\  explicitly models the probability distribution, and both forward and backward passes are meaningful.

\textbf{Unified Classification and Generation Models.} Xie \etal \cite{xie2016theory} first draw the connection between the discriminative and generative power of a ConveNet with EBMs. Based on the idea, Grathwohl \etal \cite{grathwohl2019your} propose to re-interpret a discriminative classifier as an EBM modeling joint distribution of $x$ and $y$, which enables image synthesis and classification within one framework. Introspective Neural Networks \cite{jin2017introspective, lazarow2017introspective, lee2018wasserstein} share a similar insight to imbue the classifier with generative power, leading to increased robustness in adversarial attacks. In contrast to the above methods, our method adopts the diffusion process to improve the accuracy of estimated scores for stable training and image sampling. Consequently, the regularization and training tricks required by other models are not necessary for our \OURNAME\  model. \OURNAME\  significantly outperforms the hybrid model, JEM \cite{grathwohl2019your}, by a substantial margin.

%% file: method.tex
\def\mlog{\mathrm{log}}
\def\mexp{\mathrm{exp}}

\section{Method}
\textbf{Overview.}
As illustrated in Fig.~\ref{Fig:EBDC}, the \OURNAME\ model consists of a classifier that models the joint distribution $p(\mbfx, y)$ by estimating the energy during the forward pass.
The conditional probability $p(y|\mbfx)$ is produced by Softmax function. 
The backward pass of the network produces both the unconditional score ($\nabla_\mbfx \mathrm{log}\ p(\mbfx)$) and the class guidance ($\nabla_\mbfx \mathrm{log}\ p(y|\mbfx)$) in a single step by marginalizing out $y$. We adopt the diffusion process to populate low data density regions and optimize the unconditional score using Fisher divergence, which circumvents the direct optimization of normalized probability density.

\textbf{Background.} An energy-based model \cite{lecun2006tutorial} is defined as 
\begin{equation}
    p_{\theta}(\mathbf{x}) = \frac{\mathrm{exp}(-E_{\theta}(\mathbf{x}))}{Z(\theta)},
    \label{Eq:energy}
\end{equation}
to approximate any probability density function $p_\mathrm{data}(\mathbf{x})$ for $\mathbf{x} \in \mathbb{R}^D$, where $E_{\theta}(\mathbf{x}):\mathbb{R}^D \rightarrow \mathbb{R}$, known as the energy function, maps each data point to a scalar, and $Z(\theta)=\int\mathrm{exp}(-E_\theta(\mathbf{x}))d\mathbf{x}$, analytically intractable for high-dimensional $\mathbf{x}$, is the partition function. Typically, one can parameterize the energy function with a neural network $f_\theta(\mathbf{x})=-E_\theta(\mathbf{x})$.
The {\it de facto} standard for learning probabilistic models from \textit{i.i.d} data is maximum likelihood estimation. The log-likelihood function is 
\begin{equation}
    \mathcal{L}(\theta)=\mathbb{E}_{\mathbf{x}\sim p_\mathrm{data}(\mathbf{x})}[\mathrm{log}\ p_\theta(\mathbf{x})] \simeq \frac{1}{N}\sum_{i=1}^{N}\mlog\ p_{\theta}(\mbfx_{i}),
\end{equation}
where we observe N samples $\mbfx_{i}\sim p_{\mdata}(\mbfx)$. The gradient of the log-probability of an EBM is composed of two terms:
\begin{equation}
    \frac{\partial\mlog\ p_{\theta}(\mbfx)}{\partial\theta} = \frac{\partial f_{\theta}(\mbfx)}{\partial \theta} - \mmean_{\mbfx' \sim p_\theta(\mbfx')}[\frac{\partial f_{\theta}(\mbfx')}{\partial \theta}],
\label{Eq:logp}
\end{equation}

The second term can be approximated by drawing the synthesized samples from the model distribution $p_\theta(\mbfx')$ with Markov Chain Monte Carlo (MCMC). Langevin MCMC first draws an initial sample from a simple prior distribution and iteratively updates the sample until a mode is reached, which can be formalized as follows:
\begin{equation}
    \mbfx_{i+1} \leftarrow \mbfx_{i} + c \nabla_{\mbfx}\ \mlog\ p(\mbfx_{i}) + \sqrt{2c}\bm{\epsilon}_{i},
\end{equation}
where $\mbfx_{0}$ is randomly sampled from a prior distribution (such as Gaussian distribution), and $\bm{\epsilon}_{i} \sim \mathcal{N}(\mathbf{0}, \mathbf{I})$.
However, for high-dimensional distributions, it takes a long time to run MCMC to generate a converged sample.

\subsection{Energy-Based Model with Diffusion Process}
Diffusion models gradually inject noise into the source samples during the nosing process, which can be formulated as a Markov chain:
\begin{equation}
\begin{split}
    q(\mbfx_{1:T}|\mbfx_{0}) &= \prod_{t=1}^{T}q(\mbfx_{t} | \mbfx_{t-1}),\\
    q(\mbfx_{t} | \mbfx_{t-1}) &= \mathcal{N}(\mbfx_{t}; \sqrt{\alpha_t}\mbfx_{t-1}, \beta_{t}\mathbf{I}),
\end{split}
\end{equation}
where $\mbfx_{0}$ represents the source samples and $\alpha_{t} = 1 - \beta_{t}$. For an arbitrary timestep $t$, one can directly sample from the following Gaussian distribution without iterative sampling,
\begin{equation}
q(\mbfx_{t}|\mbfx_{0}) = \mathcal{N}(\mbfx_{t}; \sqrt{\bar{\alpha}_t}\mbfx_{0}, (1-\bar{\alpha}_t)\mathbf{I})
\label{Eq:qt}
\end{equation}
Based on Bayes theorem, one can reverse the noising process by the posterior $q(\mbfx_{t-1}|\mbfx_{t}, \mbfx_{0})$:
\begin{equation}
\begin{split}
\tilde{\beta}_t &= \frac{1-\bar{\alpha}_{t-1}}{1-\bar{\alpha}_{t}} \beta_t \\
\tilde{\bm{\mu}}(\mbfx_{t}, \mbfx_{0}) &= \frac{\sqrt{\bar{\alpha}_{t-1}} \beta_t}
{1-\bar{\alpha}_t} \mbfx_0 + 
\frac{\sqrt{\alpha_{t}} (1-\bar{\alpha}_{t-1})}
{1-\bar{\alpha}_t} \mbfx_t \\
q(\mbfx_{t-1}|\mbfx_{t}, \mbfx_{0}) &= \mathcal{N}(\mbfx_{t-1}; \tilde{\bm{\mu}}(\mbfx_{t}, \mbfx_{0}), \tilde{\beta}_t\mathbf{I}) 
\end{split}
\end{equation}
$\mbfx_0$ is unknown during the denoising process, so we approximate the posterior with $p_{\theta}(\mbfx_{t-1} | \mbfx_t) = q(\mbfx_{t-1}|\mbfx_{t}, \mbfx_{0} = \bm{\mu}_{\theta}(\mbfx_t))$ to denoise the observed sample $\mbfx_t$.

According to Tweedie's Formula \cite{efron2011tweedie}, one can estimate the mean of a Gaussian distribution, given a random variable $\mathbf{z} \sim \mathcal{N}(\mathbf{z}; \bm{\mu}_{z}, \bm{\Sigma}_{z})$:
\begin{equation}
\mmean[\bm{\mu}_{z}|\mathbf{z}] = \mathbf{z} + \bm{\Sigma}_{z} \nabla_{\mathbf{z}} \mlog\ p(\mathbf{z})
\end{equation}
By applying Tweedie's Formula to Equation~\ref{Eq:qt}, the estimate for the mean of the noised sample $\mbfx_{t}$ can be represented as:
\begin{equation}
\sqrt{\bar\alpha_t}\mbfx_{0} = \mbfx_{t} + (1-\bar\alpha_t)\nabla_{\mbfx_t} \mlog\ q(\mbfx_t|\mbfx_0)
\label{Eq:tw}
\end{equation}
Since the noised sample $\mbfx_t$ decomposes as a sum of two terms: $\mbfx_t = \sqrt{\bar{\alpha}_t}\mbfx_{0} + \sqrt{1-\bar{\alpha}_t}\bm{\epsilon}_t$, the score function $\nabla_{\mbfx_t} \mlog\ q(\mbfx_t|\mbfx_0)$ can be expressed as:
\begin{equation}
\nabla_{\mbfx_t} \mlog\ q(\mbfx_t|\mbfx_0) = - \frac{\bm{\epsilon}_t}{\sqrt{1-\bar{\alpha}_t}}
\label{Eq:score}
\end{equation}
We approximate the probability density function $q(\mbfx_t|\mbfx_0)$ by an Energy-based model $p_{\theta}(\mbfx_t)$, and optimize the parameters by minimizing the Fisher divergence between $q(\mbfx_t|\mbfx_0)$ and $p_{\theta}(\mbfx_t)$:
\begin{equation}
\mathcal{D}_{F} = \mmean_q[\frac{1}{2}\| \nabla_{\mbfx_t} \mlog\ q(\mbfx_t|\mbfx_0) - \nabla_{\mbfx_t} \mlog\ p_{\theta}(\mbfx_t) \|^2]
\label{Eq:fisher}
\end{equation}
For Energy-based models, the score can be easily obtained, $\nabla_{\mbfx} \mlog\ p_{\theta}(\mbfx) = \nabla_{\mbfx} f_{\theta}(\mbfx)$. Compared with directly optimizing the log-probability of EBM (Equation~\ref{Eq:logp}), Fisher divergence circumvents optimizing the normalized densities parameterized by $Z(\theta)$ and the target score can be directly sampled from a Gaussian distribution via Equation~\ref{Eq:score}.

\subsection{\OURNAME}
\label{Sec:ECG}
The energy-based model has an inherent connection with discriminative models. For the classification problem with $C$ classes, a discriminative neural classifier maps a data sample $\mbfx \in \mathbb{R}^D$ to a vector of length $C$ known as logits. The probability of $y$-th label is represented using the Softmax function:
\begin{equation}
p(y | \mbfx) = \frac{\mexp(f(\mbfx)[y])}{\sum_{y'}\mexp(f(\mbfx)[y'])},
\label{Eq:softmax}
\end{equation}
where $f(\mbfx)[y]$ is the $y$-th logit. Using Bayes theorem, the discriminative conditional probability can be expressed as $p(y | \mbfx) = \frac{p(\mbfx, y)}{\sum_{y'}p(\mbfx, y')}$. By connecting Equation~\ref{Eq:energy} and \ref{Eq:softmax}, the joint probability of data sample $\mbfx$ and label $y$ can be modeled as:
\begin{equation}
p_{\theta}(\mbfx, y) = \frac{\mexp(f_{\theta}(\mbfx)[y])}{Z(\theta)}
\end{equation}
By marginalizing out $y$, the score of data sample $\mbfx$ is obtained as:
\begin{equation}
\nabla_{\mbfx} \mlog\ p_{\theta}(\mbfx) = \nabla_{\mbfx} \mlog \sum_{y} \mexp(f_{\theta}(\mbfx)[y])
\label{Eq:logsumexp}
\end{equation}
Different from the typical classifiers, the free energy function $E_{\theta}(\mbfx) = - \mlog \sum_{y} \mexp(f_{\theta}(\mbfx)[y])$ is also optimized for generating samples. 

We propose to integrate the energy-based classifier with the diffusion process to achieve both strong discriminative performance and generative performance. 
Specifically, we approximate the conditional probability density function $q(\mbfx_t, y|\mbfx_0)$ with an energy-based classifier $p_{\theta}(\mbfx_t, y)$. Due to the optimization of Fisher divergence for our EBM, we factorize the log-likehood as:
\begin{equation}
\mlog\ p_{\theta}(\mbfx_t, y) = \mlog\ p_{\theta}(\mbfx_t) + \mlog\ p_{\theta}(y | \mbfx_t)
\end{equation}
The score $\nabla_{\mbfx_t} \mlog\ p_{\theta}(\mbfx_t)$ is optimized by minimizing the Fisher divergence as shown in the Equation~\ref{Eq:fisher} and \ref{Eq:logsumexp}. As for the conditional probability $p_{\theta}(y | \mbfx_t)$, we simply adopt the standard cross-entropy loss to optimize.

One of the advantage of integrating the energy-based classifier with diffusion process is that the classifier provides guidance to explicitly control the data we generate through conditioning information $y$. By Bayes theorem, the conditional score can be derived as:
\begin{equation}
\nabla \mlog\ p_{\theta}(\mbfx_t | y) = \nabla \mlog\ p_{\theta}(\mbfx_t) + \nabla \mlog\ p_{\theta}(y | \mbfx_t)
\end{equation}

The joint probability $p_{\theta}(\mbfx_t, y)$ is parameterized with a neural network. And the forward propagation of our \OURNAME\ model is a discrimination model to predict the conditional probability $p_{\theta}(y | \mbfx)$, while the backward propagation of the neural network is a generation model to predict the score and classifier guidance to gradually denoise data.

Overall, the training loss of an \OURNAME\ model is formulated as:
\begin{equation}
\begin{split}
\mathcal{L} = &\ \mmean_q[\frac{1}{2}\| \nabla_{\mbfx_t} \mlog\ q(\mbfx_t|\mbfx_0) - \nabla_{\mbfx_t} \mlog\ p_{\theta}(\mbfx_t) \|^2 \\
            & - \sum_{i=1}^{C} q(y_i|\mbfx_t, \mbfx_0)\ \mlog\ p_{\theta}(y_i | \mbfx_t)],
\end{split}
\end{equation}
where the first term is reconstruction loss for a noised sample, the second term is a classification loss that encourages the denoising process to generate samples that are consistent with the given labels. The training procedure is summarized in Algorithm~\ref{algo:1}, where we adopt the noise $\bm{\epsilon}$ as the target score to ensure the stable optimization of the neural network. Additionally, a hyperparameter $\gamma$ is introduced to balance the two loss terms.

\begin{algorithm}[H]
\caption{Training}
\label{algo:1}	
\begin{algorithmic}
	\REPEAT
	\STATE Sample $t \sim {\rm Unif}(\{1, ..., T\})$
	\STATE Sample data pair $(\mbfx_{0}, y)$, Sample noise $\bm{\epsilon} \sim \mathcal{N}(\mathbf{0}, \mathbf{I})$
        \STATE $\mbfx_t = \sqrt{\bar{\alpha}_t} \mbfx_{0} + \sqrt{1 - \bar{\alpha}_t} \bm{\epsilon}$ 
        \STATE Take gradient descent step on 
        $\nabla_{\theta} (\| \nabla_{\mbfx_t} \mlog\ p_{\theta}(\mbfx_t) + \bm{\epsilon} \|^2 - \gamma \sum_{i=1}^{C} q(y_i|\mbfx_t)\ \mlog\ p_{\theta}(y_i | \mbfx_t) )$ 
	\UNTIL converged.
\end{algorithmic}
\end{algorithm}

%% file: exp.tex
\section{Experiments}
We conduct a series of experiments to evaluate the performance of \OURNAME\ model on image classification and generation benchmarks. The results show that our model achieves performance rivaling the state of the art in both discriminative and generative modeling.

\input{tables/cifar10_hybrid.tex}

\subsection{Experimental Setup}
For conditional learning, we consider CIFAR-10 \cite{krizhevsky2009learning}, CIFAR-100 and ImageNet \cite{deng2009imagenet} dataset
to evaluate the proposed \OURNAME\ model. Both CIFAR-10 and CIFAR-100 contain 50K training images.
The ImageNet training set is composed of about 1.28 million images from 1000 different categories. 
For unconditional learning, we train unsupervised \OURNAME\ models on CelebA-HQ \cite{karras2017progressive}, which contains 30K training images of human faces, and LSUN Church \cite{yu2015lsun}, which contains about 125K images of outdoor churches. For ImageNet-1k, CelebA-HQ and LSUN Church, we follow latent diffusion models (LDM) \cite{rombach2022high} to convert images at 256$\times$256 resolution to latent representations at 32$\times$32 or 64$\times$64 resolutions, using the pre-trained image autoencoder provided by LDM \cite{rombach2022high}. We adopt the same UNet architecture as \cite{dhariwal2021diffusion} and attach an attention pooling module to it, like CLIP \cite{radford2021learning}, to predict logits. More training details can be found in Appendix.

\subsection{Hybrid Modeling}

\textbf{\OURNAME\ model.}
We first train an \OURNAME\ model on CIFAR-10 dataset. The accuracy on CIFAR-10 validation dataset, Inception Score (IS) \cite{salimans2016improved} and Frechet Inception Distance (FID) \cite{heusel2017gans} are reported to quantify the performance of our model. 
Our model demonstrates a remarkable accuracy of 95.9\% on the CIFAR-10 validation dataset, surpassing the performance of the discriminative model Wide ResNet-28-12, which has a comparable architecture and number of parameters. Furthermore, the sampling quality of our model outperforms a majority of existing Explicit EBM, GAN models and Score-Based models, with IS of 9.43 and FID of 3.30. 
These results showcase the potential of our proposed model to enhance image classification performance and generative capability of EBM.
On CIFAR-100, our proposed \OURNAME\ model achieves comparable generative results with state-of-the-art GAN-based methods, while outperforming the hybrid model JEM \cite{grathwohl2019your} in terms of classification accuracy, as shown in Table~\ref{tab:CIFAR_100}.

We conducted additional experiments on the more challenging dataset, ImageNet. The results, presented in Table~\ref{tab:IMGNET}, provide evidence that our proposed \OURNAME\ model performs well on this dataset, achieving IS of \IMGNETIS\, FID of \IMGNETFID\ and Top-1 Accuracy of\ \IMGNETACC \%. The model was trained using only random flip as data augmentation. We recognize that incorporating stronger augmentation techniques could lead to even better results. By merely incorporating the \textit{RandResizeCrop} data augmentation, we achieve a significant increase of\ \IMGNETWAACC\% in accuracy on the ImageNet dataset.

\input{tables/cifar100.tex}

\input{tables/imgnet.tex}

\textbf{\UCNAME.} 
\OURNAME\  model can be trained in an unsupervised
manner without knowing the label of image. By marginalizing out the variable $y$ in Equation~\ref{Eq:logsumexp}, the score function can be optimized by minimizing the Fisher divergence.  
As shown in Table~\ref{tab:MultiDataSet}, \UCNAME\ achieves FID of \CELEHQRES\ on CelebA-HQ 256$\times$256 and FID of \CHURCHRES\ on LSUN Church 256$\times$256. These results demonstrate that \UCNAME\ outperforms the other state-of-the-art Energy-Based models. Although the score-based diffusion model, DDPM, exhibits slightly better performance, it is noteworthy that optimizing the gradient of the neural network for \UCNAME\ is a more challenging task. We believe that a network architecture specifically designed for optimizing the gradient will lead to the same or even better results than DDPM.

\input{tables/unsup.tex}
\input{tables/ablation.tex}

\textbf{Ablation study.} 
The results in Table~\ref{tab:ablation} demonstrate the effectiveness of the proposed \OURNAME\ framework for image synthesis and classification.
\UCNAME\ serves as a good baseline, achieving FID of 5.36. The \OURNAME\ model achieves 95.9\% accuracy on the test set and an additional 1.87 FID improvement, demonstrating the success of learning the joint probability. Moreover, we make use of class labels for conditional image synthesis. The classifier guidance $\nabla \mlog\ p_{\theta}(y | \mbfx_t)$ guides the denoise process towards the class label $y$, resulting in a 0.19 FID improvement. To investigate the effect of neural network architecture, we train a model based on the standard feedforward ResNet commonly used for image classification. The comparison of the last and penultimate lines in Table~\ref{tab:ablation} reveals that the U-Net architecture benefits from the short-cut connections specifically designed to propagate fine details from the inputs x.

\subsection{Application and Analysis}

\begin{figure}
\begin{center}
    \includegraphics[width=\columnwidth]{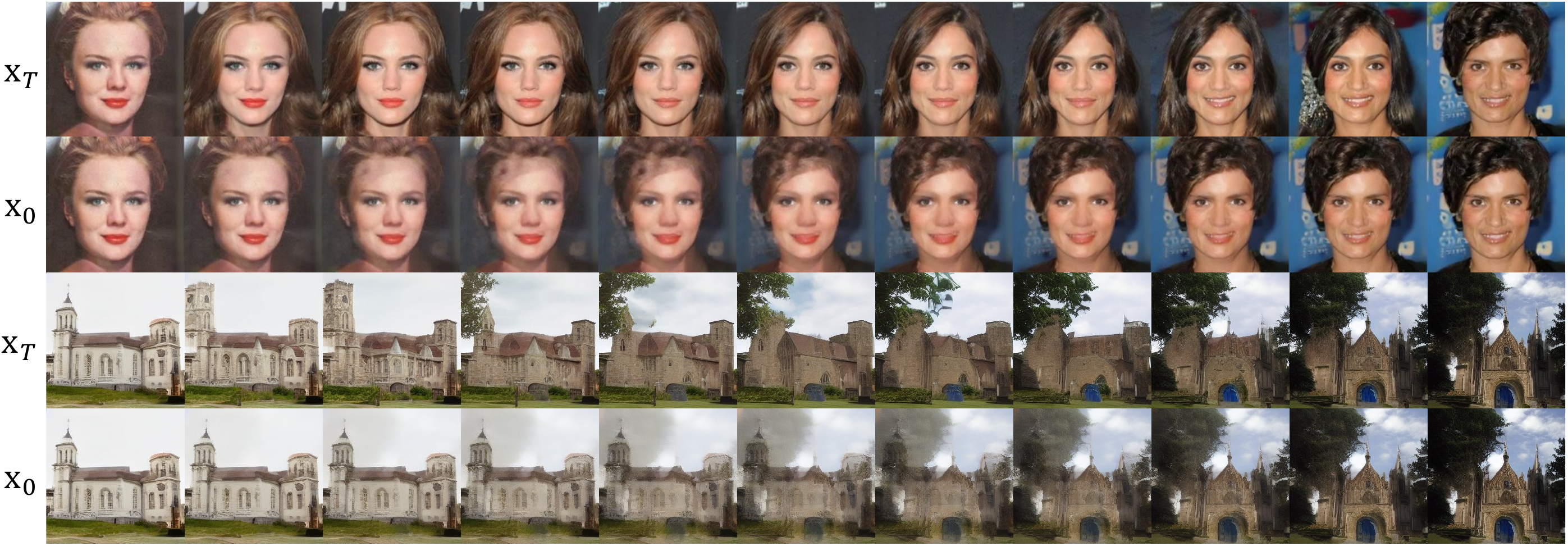}
\end{center}
\vspace{-10pt}
\caption{Interpolation results between the leftmost and rightmost generated samples. $\mbfx_{T}$ denotes the interpolation the noise samples. $\mbfx_{0}$ means the interpolation on the generated samples. The results demonstrate that our proposed method exhibits superior semantic interpolation effects compared to direct interpolation of generated samples in the latent space.}
\label{Fig:interpolate}
\vspace{-5pt}
\end{figure}

\textbf{Interpolation.} 
As shown in Figure~\ref{Fig:interpolate}, our model is capable of producing smooth interpolation between two generated samples. Following DDIM \cite{song2020denoising}, we perform an interpolation between the initial white noise samples $\mbfx_{T}$. We compare the generated samples from the interpolated noise with the samples obtained through interpolation in the generated samples $\mbfx_{0}$. The results demonstrate that our method exhibits superior semantic interpolation effects compared to direct interpolation of generated samples in latent space.

\begin{figure}
\begin{center}
    \includegraphics[width=\columnwidth]{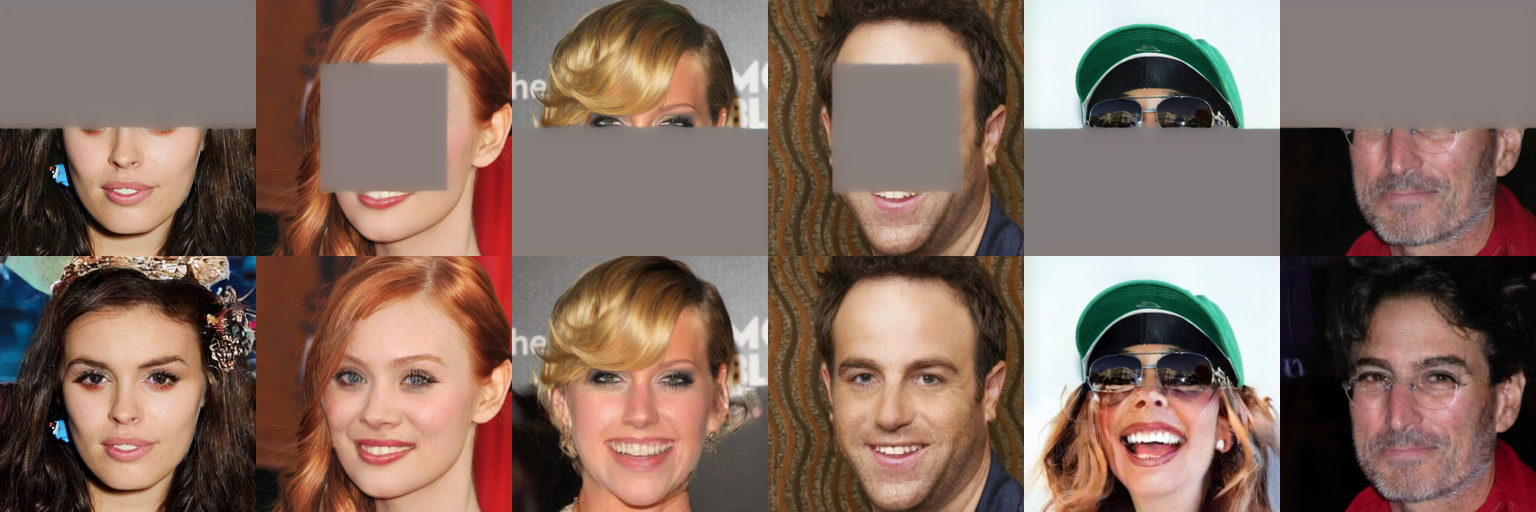}
\end{center}
\vspace{-10pt}
\caption{Inpainting results on CelebA-HQ dataset with resolution of 256$\times$256. The top row shows the mask images, while the bottom row displays the corresponding inpainted images.}
\label{Fig:inpaint}
\vspace{-10pt}
\end{figure}

\textbf{Image inpainting.} 
A promising application of energy-based models is to use the learned prior model for filling masked regions of an image with new content. 
Following \cite{lugmayr2022repaint}, we obtain a sequence of masked noisey  image at different timesteps and fill the masked pixels with the denoised sample given the previous iteration. The qualitative results on CelebA-HQ 256$\times$256 are presented in Figure~\ref{Fig:inpaint}, demonstrating that our model is capable of realistic and semantically meaningful image inpainting.

\begin{figure*}
     \centering
     \begin{subfigure}[b]{0.32\textwidth}
         \centering
         \includegraphics[trim=1cm 0cm 0cm 0cm,width=\columnwidth]{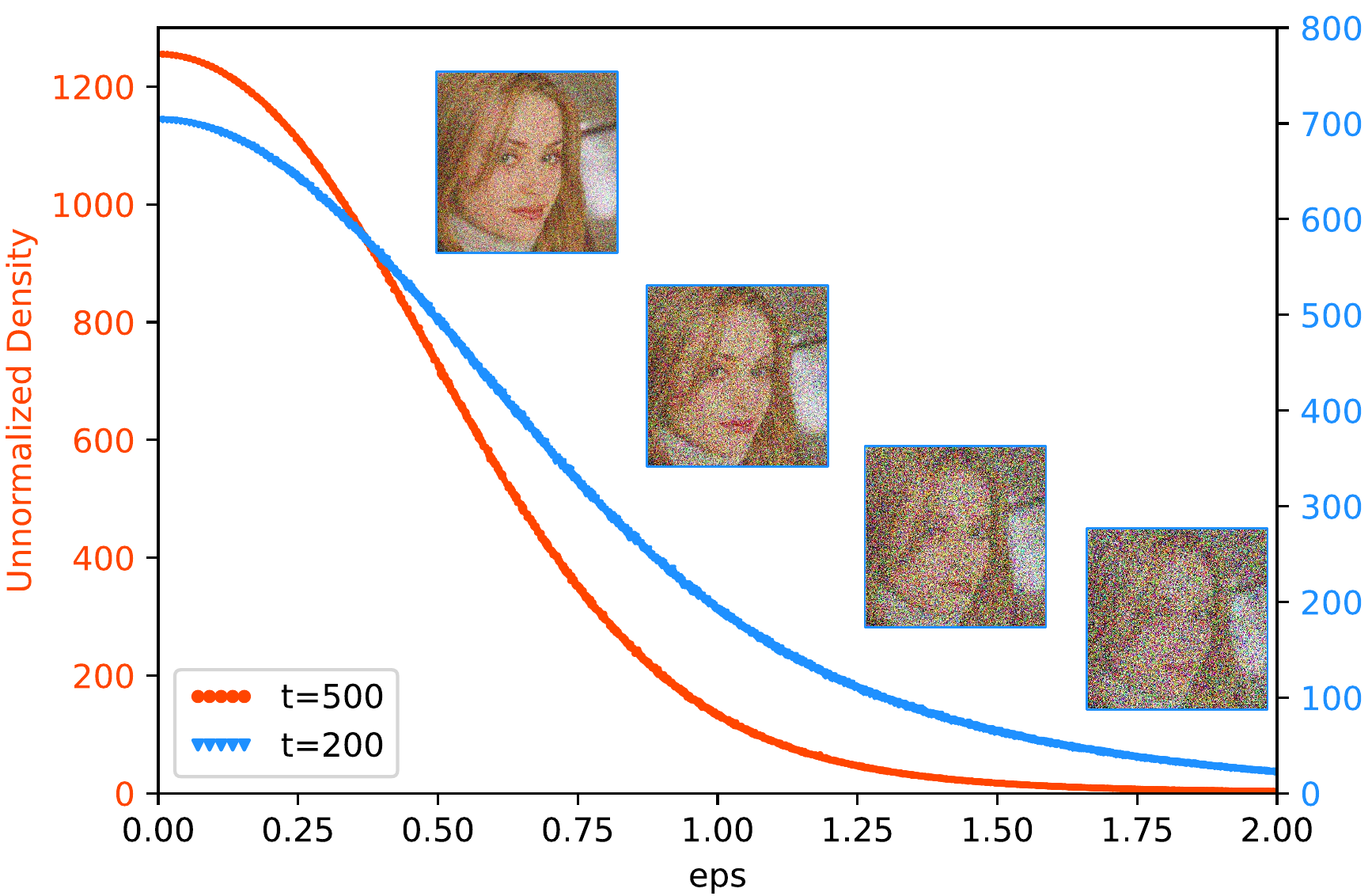}
         \caption{}
         \label{Fig:density}
     \end{subfigure}
     \hfill
     \begin{subfigure}[b]{0.32\textwidth}
         \centering
         \includegraphics[trim=0.5cm 0cm 1.1cm 0cm,width=\columnwidth]{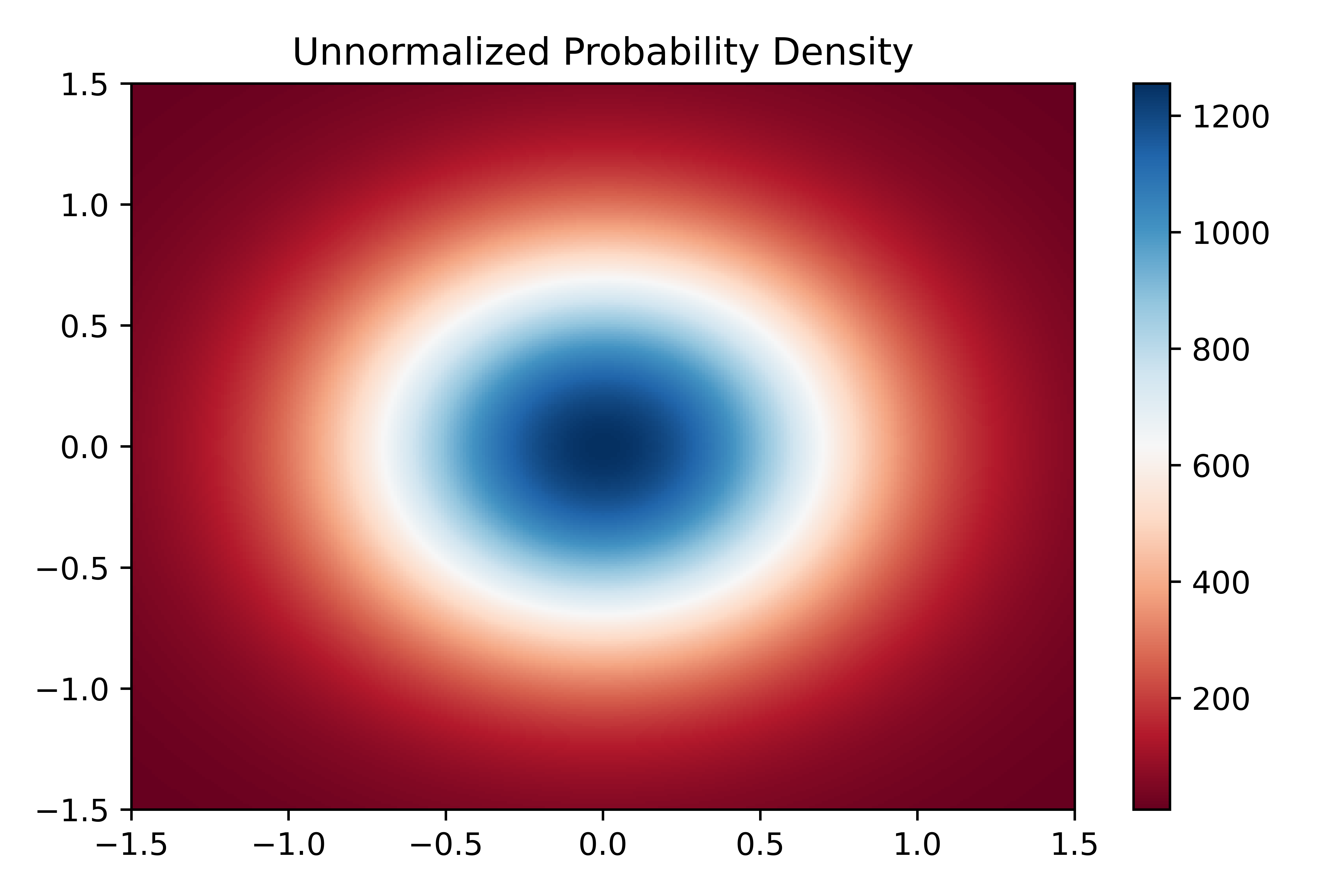}
         \caption{}
         \label{Fig:density_2d}
     \end{subfigure}
     \hfill
     \begin{subfigure}[b]{0.32\textwidth}
         \centering
         \includegraphics[trim=0.5cm 0cm 1.1cm 0cm, width=\columnwidth]{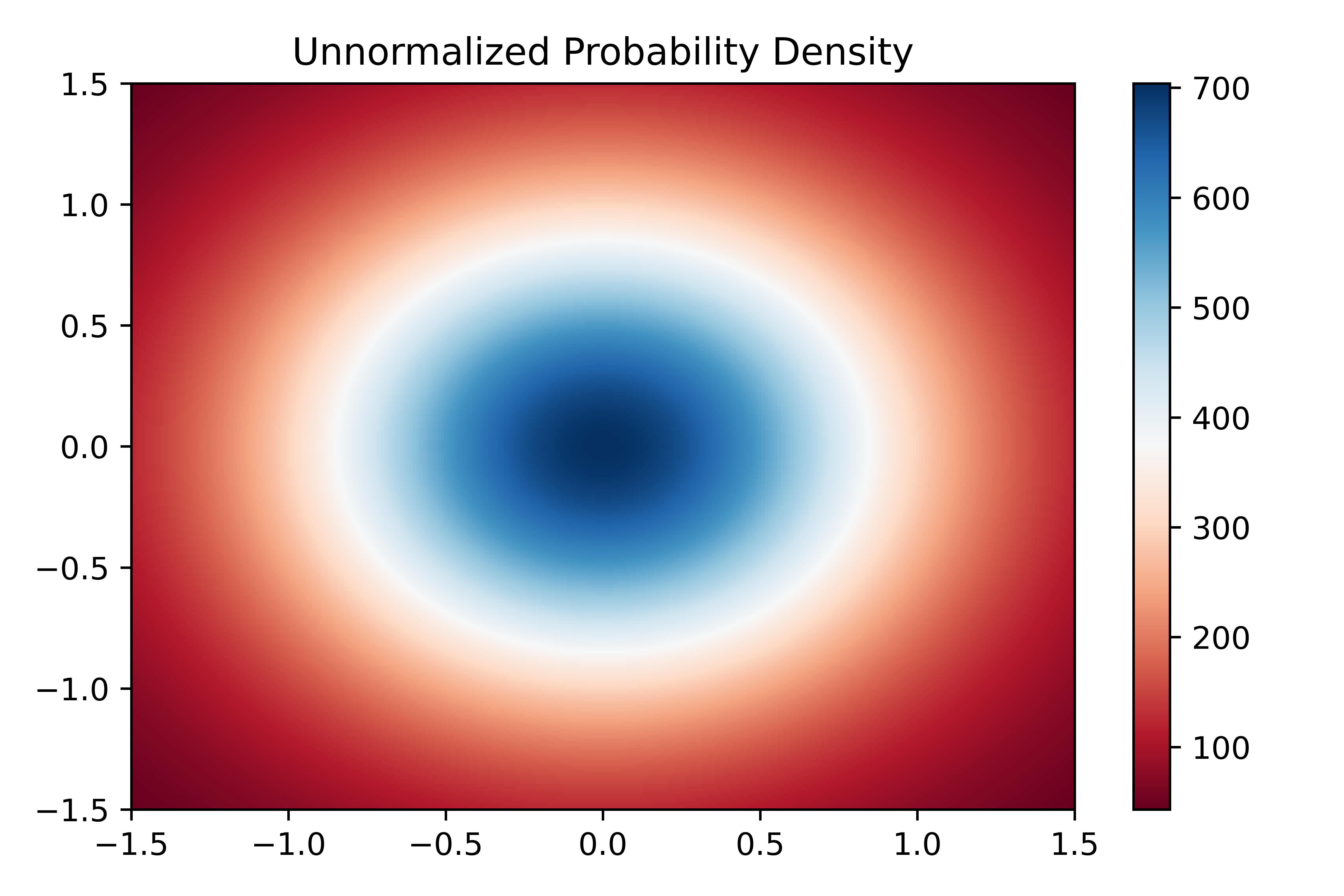}
         \caption{}
     \end{subfigure}
        \vspace{-5pt}
        \caption{(a) The unnormalized probability density. The x-axis represents the noise level of $\mbfx_{t}$, which is the mean of the abstract value of the noise at each timestep. The probability density exhibits a similar shape to the folded normal distribution. (b) The density function of the noised samples at $t=500$. The noise is produced by linearly combining two orthogonal noises. The probability density exhibits a similar shape to the Gaussian distribution. (c) The density function of the noised samples at $t=200$.}
        \label{fig:three graphs}
        \vspace{-5pt}
\end{figure*}

\begin{figure}
\begin{center}
    \includegraphics[trim=0cm 0cm 0cm 0.45cm,width=\columnwidth]{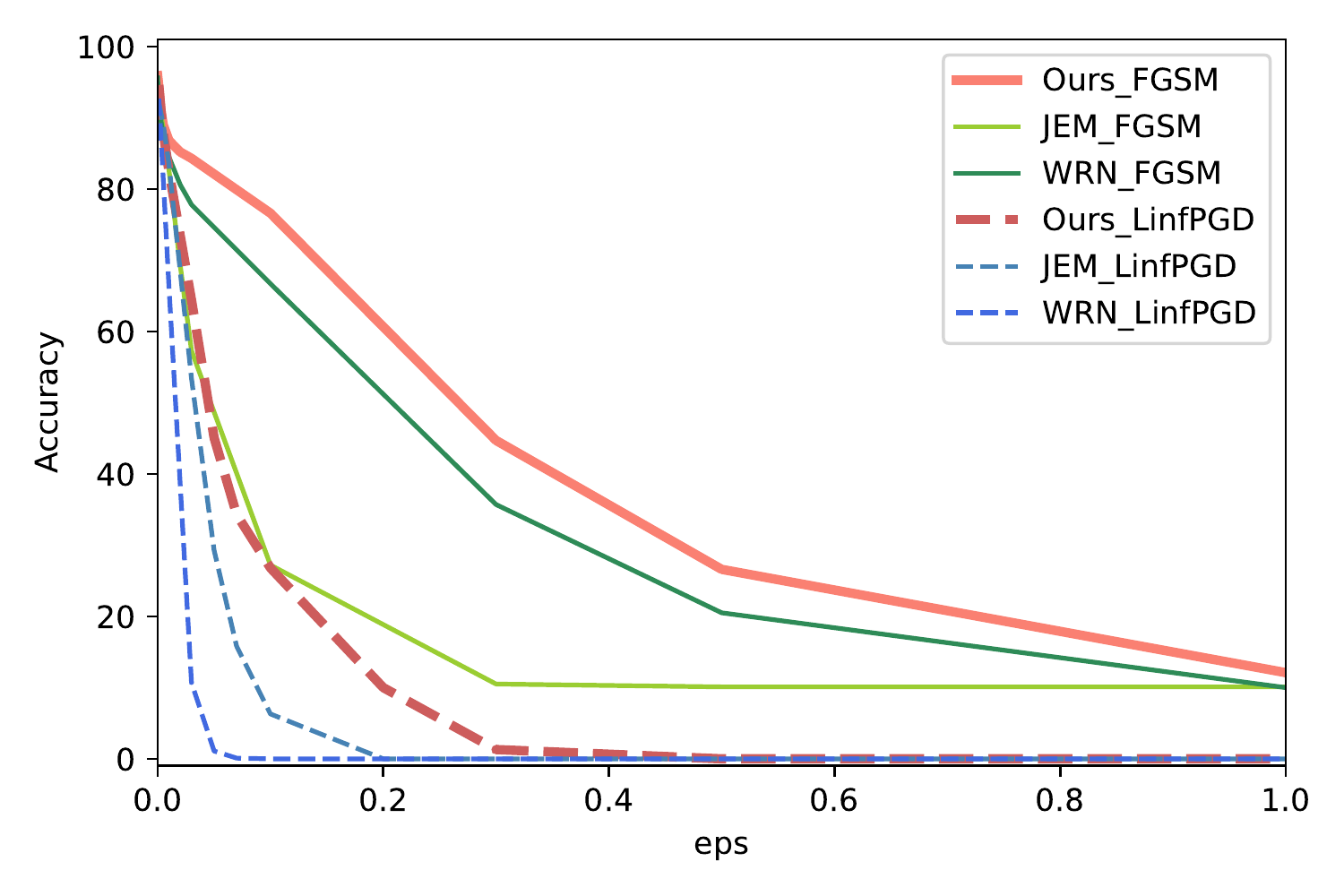}
\end{center}
\vspace{-22pt}
\caption{Robustness evaluation of \OURNAME\ model on CIFAR-10 with PGD and FGSM adversarial attacks. Our proposed classifier exhibits considerable improvement in adversarial robustness compared to the baseline.}
\label{Fig:robustness}
\vspace{-10pt}
\end{figure}

\textbf{Robustness.} 
Adversarial attacks are a common threat to neural networks, especially when the model weights are accessible. A common white-box attack is FGSM, perturbing the inputs with the gradients enlarging the loss. We investigate the robustness of models trained on the CIFAR-10 dataset using FGSM \cite{goodfellow2014explaining} and PGD \cite{madry2017towards} attacks under the $L_{\infty}$ constraint. The accuracy curve in Figure~\ref{Fig:robustness} demonstrates that the \OURNAME\ model outperforms the standard Wide ResNet-28-12 classifier and JEM \cite{grathwohl2019your} in terms of adversarial robustness. The results suggest that leveraging the joint probability distribution learned by energy-based models can enhance the model's robustness against adversarial attacks.

\textbf{Visualize Energy.} 
As detailed in Section~\ref{Sec:ECG}, we adopt a direct optimization of the Fisher divergence instead of the probability $p_\theta(\mbfx_t)$. Therefore, we are interested to see whether the neural network would effectively model the target Gaussian distribution. Given the difficulty in illustrating a high-dimensional Gaussian distribution, we present that the unnormalized probability density of noised sample $\mbfx_t$ in Figure~\ref{Fig:density}.
Notably, the density exhibits a similar shape to the folded normal distribution, suggesting that the probability distribution learned by the neural network closely approximates the Gaussian distribution. 
In Figure~\ref{Fig:density_2d}, we select two orthogonal noises to plot the two-dimensional probability density function, which exhibits a similar shape to the Gaussian distribution.

\begin{figure}[t]
\begin{center}
    \includegraphics[trim=0cm 0cm 0cm 0cm,width=\columnwidth]{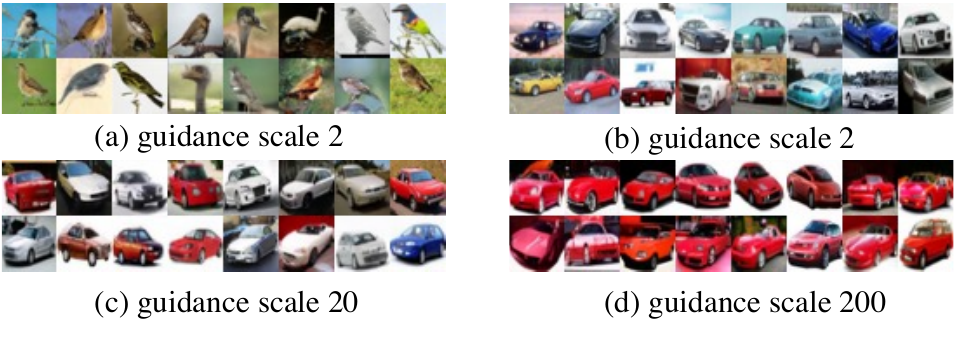}
\end{center}
\vspace{-10pt}
\caption{The conditional sampling results demonstrate that the random noise promises the diversity and the classification guidance guarantees the semantic consistency. As the guidance scale increase, the diversity significantly decreases and the mode collapse to the class prototype.}
\label{Fig:condsample}
\end{figure}

\textbf{Conditional sampling.} 
As illustrated in Section~\ref{Sec:ECG}, the classifier provides guidance to explicitly control the data we generate
through conditioning information $y$. We feed the network with a fixed class label and random noise to check the qualitative results. As shown in Fig.~\ref{Fig:condsample}, the diversity is promised by the random noise and the semantic consistency is guaranteed by the classification guidance. We further investigate the relationship between  guidance scale with sample diversity. By increasing the guidance scale from 2 to 200, the diversity decreases significantly. An interesting observation is that the mode finally collapse to the class prototype. Meanwhile, the generated image with a high guidance scale are of high fidelity.

\section{Conclusion}
In this work, we introduce a novel energy-based model, \OURNAME, to bridge the gap between discrimative and generative learning. We formulate the image classification and generation tasks from a probabilistic perspective, and find that the energy-based model is suitable for both tasks. To alleviate the difficult of optimizing the normalized probability density of energy-based model, we introduce diffusion process to populate the low data density regions for better estimated score, and score matching to circumvent optimizing the object loss with the normalizing constant $Z(\theta)$. In \OURNAME, the forward pass models the joint distribution of the noisy image and label, while the backward pass of \OURNAME\  calculates both conditional and unconditional scores to encourage the denoising process to generate samples consistent with the given labels. We achieve high-quality image synthesis and competitive image classification accuracy using a single neural network. We believe that \OURNAME\  can serve as a new baseline for unifying discriminative and generative models.

%% file: tables/cifar10_hybrid.tex
\begin{table}[t]
\begin{center}
\begin{small}
\begin{tabular}{lcccc}
\toprule
\textbf{Method} & Acc(\%) & IS($\uparrow$) & FID($\downarrow$) \\
\midrule
\textit{Hybrid Model} &&& \\
\midrule
Ours & \textbf{95.9} & \textbf{9.43} & \textbf{3.30} \\
Glow \cite{kingma2018glow} & 67.6 & 3.92 & 48.9 \\
R-Flow \cite{chen2019residual} & 70.3 & 3.60 & 46.4 \\
IGEBM \cite{du2019implicit} & 49.1 & 8.30 & 37.9 \\
JEM \cite{grathwohl2019your} & 92.9 & 8.76 & 38.4 \\
\midrule
\textit{Explicit EBM} &&& \\
\midrule
 Diff Recovery \cite{gao2020learning} & N/A & 8.30 & 9.58 \\
 VAEBM \cite{xiao2020vaebm} & N/A & 8.43 & 12.2 \\
 ImprovedCD \cite{du2020improved} & N/A & 7.85 & 25.1 \\
 CF-EBM \cite{zhao2020learning} & N/A & - & 16.7 \\
 EBMs-VAE \cite{xie2021learning} & N/A & 6.65 & 36.2 \\
 CoopFlow \cite{xie2022tale} & N/A & - & 15.8 \\
 CEM \cite{wang2022unified} & N/A & 8.68 & 36.4 \\
 ATEBM \cite{yin2022learning} & N/A & 9.10 & 13.2 \\
 HATEBM \cite{hill2022learning} & N/A & - & 19.30 \\
 Adaptive CE \cite{xiao2022adaptive} & N/A & - & 65.01 \\
 CLEL \cite{leeguiding} & N/A & - & 8.61 \\
\midrule
\textit{GANs} &&& \\
\midrule
 BigGAN \cite{brock2018large} & N/A & 9.22 & 14.7 \\
 SNGAN \cite{miyato2018spectral} & N/A & 8.22 & 21.7 \\
 StyleGAN* \cite{karras2020training} & N/A & 8.99 & 9.90 \\
 \midrule
\textit{Score-Based Model} &&& \\
 \midrule
 DDPM \cite{ho2020denoising} & N/A & \textbf{9.46} & \textbf{3.17} \\
 NCSN \cite{song2019generative} & N/A & 8.87 & 25.32 \\
 NCSN-v2 \cite{song2020improved} & N/A & 8.40 & 10.87 \\
 \midrule
 \textit{Discriminative Model} &&& \\
 \midrule
 Wide ResNet-28-12 \cite{zagoruyko2016wide} & 95.6 & N/A & N/A \\
\bottomrule
\end{tabular}
\end{small}
\end{center}
\vspace{-10pt}
\caption{CIFAR-10 hybrid modeling results. `N/A' means the corresponding result is not available. We report the result of \textit{Wide ResNet-28-12}, which has a similar architecture, number of parameters and computational cost to our proposed model.}
\label{tab:C10Hybrid}
\vspace{-10pt}
\end{table}

%% file: tables/cifar100.tex
\begin{table}[t]
\begin{center}
\begin{small}
\begin{tabular}{lcccc}
\toprule
\textbf{Method} & Acc(\%) & IS($\uparrow$) & FID($\downarrow$) \\
\midrule
Ours & 77.9 & \textbf{11.50} & \textbf{4.88} \\
FQ-GAN \cite{zhao2020feature} & N/A & 7.15 & 9.74 \\
LeCAM (BigGAN) \cite{tseng2021regularizing} & N/A & - & 11.2 \\
StyleGAN2 + DA \cite{karras2020analyzing} & N/A & - & 15.22 \\
JEM \cite{grathwohl2019your} & 72.2 & - & - \\
Wide ResNet-28-12 \cite{zagoruyko2016wide} & 79.5 & N/A & N/A \\
\bottomrule
\end{tabular}
\end{small}
\end{center}
\vspace{-10pt}
\caption{Results of \OURNAME\ model on CIFAR-100 dataset. Our model outperforms other state-of-the-art generative models in terms of FID and IS, while achieving superior classification accuracy.}
\label{tab:CIFAR_100}
\end{table}

%% file: tables/imgnet.tex
\begin{table}[t]
\begin{center}
\begin{small}
\begin{tabular}{lcccc}
\toprule
\textbf{Method} & Acc(\%) & IS($\uparrow$) & FID($\downarrow$) \\
\midrule
\OURNAME$^\ddag$ & \textbf{\IMGNETWAACC} & \textbf{\IMGNETWAIS} & \textbf{\IMGNETWAFID} \\
\OURNAME$^\dag$ & \IMGNETACC & \IMGNETIS & \IMGNETFID \\
\OURNAME & 70.4 & 79.9 & 17.5 \\
ADM-Classifier \cite{dhariwal2021diffusion} & 64.3 & N/A & N/A \\
HATEBM \cite{hill2022learning} (128$\times$128) & N/A & - & 29.37 \\
IGEBM \cite{du2019implicit} (128$\times$128) & N/A & 28.6 & 43.7 \\
IDDPM \cite{nichol2021improved} & N/A & - & 12.3 \\
ADM \cite{dhariwal2021diffusion} & N/A & 100.98 & 10.94 \\
LDM-VQ-8 \cite{rombach2022high} & N/A & 201.56 & 7.77 \\
VQGAN \cite{esser2021taming} & N/A & 78.3 & 15.78 \\
\bottomrule
\end{tabular}
\end{small}
\end{center}
\vspace{-10pt}
\caption{Results of \OURNAME\ model on ImageNet-1k 256$\times$256 dataset using only random flip as data augmentation. $^\dag$ represents jointly training a conditional and an unconditional model. $^\ddag$ represents incorporating the \textit{RandResizeCrop} data augmentation.
We believe that a stronger augmentation strategy would likely yield improved results.}
\label{tab:IMGNET}
\vspace{-10pt}
\end{table}

%% file: tables/unsup.tex





\begin{table}[t]
\begin{minipage}{0.495\linewidth}
\centering
\resizebox{\linewidth}{!}{
\begin{tabular}{lc}
\toprule
\textbf{CelebA-HQ 256$\times$256} &  FID($\downarrow$) \\
\midrule
Ours & \textbf{\CELEHQRES} \\
ATEBM \cite{yin2022learning} & 17.31 \\
VAEBM \cite{xiao2020vaebm} & 20.38 \\
CF-EBM \cite{zhao2020learning} (128$\times$128) & 23.50 \\
NVAE \cite{vahdat2020nvae} & 45.11 \\
Glow \cite{kingma2018glow} & 68.93 \\
ProgressiveGAN \cite{karras2017progressive} & 8.03 \\
\bottomrule
\end{tabular}
}
\end{minipage}
\hfill
\begin{minipage}{0.495\linewidth}
\centering
\resizebox{\linewidth}{!}{
\begin{tabular}{lc}
\toprule
\textbf{LSUN Church 256$\times$256} & FID($\downarrow$) \\
\midrule

Ours & \textbf{\CHURCHRES} \\
VAEBM \cite{xiao2020vaebm} (64$\times$64) & 13.51 \\
ATEBM \cite{yin2022learning} & 14.87 \\
DDPM \cite{ho2020denoising} & 7.89 \\
\bottomrule
\end{tabular}
}
\end{minipage}

\vspace{-3pt}
\caption{Results of \UCNAME s on the CelebA-HQ and LSUN Church datasets. Our model outperforms the other state-of-the-art Energy-Based Models in terms of FID.}
\label{tab:MultiDataSet}

\end{table}

%% file: tables/ablation.tex
\begin{table}[t]
\begin{center}
\begin{small}
\resizebox{\linewidth}{!}{
\begin{tabular}{cccccc}
\toprule
EBM & Classifier & Guidance & Network & Acc(\%) & FID($\downarrow$) \\
\midrule
\checkmark & & & U-Net & N/A & 5.36 \\
\checkmark & \checkmark & & U-Net & \textbf{95.9} & 3.49 \\
\checkmark & \checkmark & \checkmark & U-Net & \textbf{95.9} & \textbf{3.30} \\
\checkmark & \checkmark & \checkmark & ResNet & \textbf{95.9} & 7.15 \\
\bottomrule
\end{tabular}
}
\end{small}
\end{center}
\vspace{-10pt}
\caption{The ablative results on CIFAR-10 dataset. EBM refers to \UCNAME\ model. Classifier represents \OURNAME, and Guidance indicates the use of the gradient of the label for generating samples. Additionally, we 
evaluate the effectiveness of our method using an energy-based model based on a standard feedforward ResNet as often used for image classification.}
\label{tab:ablation}
\vspace{-10pt}
\end{table}

%% file: appendix.tex
\appendix
\section{Appendix}

\subsection{Implementation Details}

\begin{algorithm}[H]
\caption{Sampling}
\label{algo:2}	
\begin{algorithmic}
\STATE Sample $\mbfx_{T} \sim \mathcal{N}(\mathbf{0}, \mathbf{I})$
\STATE \textbf{for} $t = T,...,1$ \textbf{do}
\STATE \ \ \ \ \ \ Sample noise $\bm{\epsilon} \sim \mathcal{N}(\mathbf{0}, \mathbf{I})$ if $t > 1$, else $\bm{\epsilon} = 0$
\STATE \ \ \ \ \ \ $\mbfx_{t-1} = \frac{1}{\sqrt{\alpha_t}}(\mbfx_t + (1-\alpha_t)\nabla_{\mbfx_t} \mathrm{log}\ p_{\theta}(\mbfx_t)) + \sqrt{\beta_t}\bm{\epsilon}$
\STATE \textbf{end for}
\STATE \textbf{return} $\mbfx_{0}$
\end{algorithmic}
\end{algorithm}

We adopt the UNet architecture used in LDM \cite{rombach2022high} and IDDPM \cite{nichol2021improved}, with the group normalization layers retained. To improve convergence speed, we do not include spectral normalization and weight normalization regularization. We adjust the channel width, multiplier, attention resolution, and depth compared to IDDPM and LDM, as shown in Tab.~\ref{tab:structure}. We use the {\it conv\_resample} instead of the {\it ResDown/Up Block} to upsample and downsample features. To optimize our model, we set the learning rate to 0.0001, batch size to 128 and weight decay to 0 across all datasets except ImageNet, for which we use a batch size of 512. The Adam optimizer is used to update the model parameters. To save computational resources, the ImageNet and LSUN Church images are compressed to 32$\times$32$\times$4 latent features by the KL-autoencoder \cite{rombach2022high}, while CelebA-HQ images are compressed to 64$\times$64$\times$3 latent features. To augment the data, we randomly flip the images for all the datasets except CIFAR, for which all the images for training are padded with 4 pixels on each side and a 32 × 32 crop is randomly sampled from the padded image or its horizontal flip, and cutout is used to avoid overfitting. The classification results at $t=0$ are reported in Table~1-3. To balance the reconstruction and classification losses, we set $\gamma=0.001$ for CIFAR and $\gamma=0.005$ for ImageNet in Algorithm.~1.

We follow the sampling strategy used in DDPM and describe it in detail in Algo.~\ref{algo:2}. To conduct conditional sampling, we replace the unconditional score $\nabla_{\mbfx_t} \mathrm{log}\ p_{\theta}(\mbfx_t)$ with the conditional score $\nabla_{\mbfx_t} \mathrm{log}\ p_{\theta}(\mbfx_t | y)$.

\input{tables/struct.tex}

%% file: tables/struct.tex
\begin{table}[h]
\begin{center}
\resizebox{\linewidth}{!}{
\begin{tabular}{lcccc}
\toprule
    & CIFAR & LSUN-Church & CelebA-HQ & ImageNet \\
\midrule
Diffusion steps & 1000 & 1000 & 1000 & 1000 \\
Noise Schedule & cosine & linear & linear & linear \\
Channel & 192 & 256 & 256 & 384 \\
Depth & 3 & 2 & 3 & 2 \\
Channel Multiplier & 1,2,2 & 1,2,2 & 1,2,3,4 & 1,2,4 \\
Attention Resolution & 16,8 & 32,16,8 & 16,8 & 32,16,8 \\
Iteration & 200k & 200k & 100k & 700k \\
Batch Size & 128 & 128 & 128 & 512 \\
\bottomrule
\end{tabular}
}
\end{center}
\vspace{-10pt}
\caption{The hyperparameters of the \OURNAME\ models producing the results shown in Table~1-3.}
\label{tab:structure}
\end{table}